\documentclass[runningheads]{llncs}
\usepackage[T1]{fontenc}
\usepackage{multirow}
\usepackage{epsfig}
\usepackage{graphicx}
\usepackage{amsmath}
\usepackage{amssymb}
\usepackage{cite}
\begin{document}
\title{3D Face Reconstruction With Geometry Details
    From a Single Color Image Under Occluded Scenes}
%
%
\author{Dapeng Zhao\inst{1} \and
    Yue Qi\inst{1,2,3} }
\institute{State Key Laboratory of Virtual Reality Technology and Systems,School of Computer Science and Engineering at Beihang University, Beijing, China  \\
    \email{mirror1775@gmail.com}\\
    \and
    Peng Cheng Laboratory, Shenzhen, China\\
    \and
    Qingdao Research Institute of Beihang University, Qingdao, China \\
    \email{}
}
\maketitle              
\begin{abstract}
    3D face reconstruction technology aims to generate a face stereo model naturally and realistically. Previous deep face reconstruction approaches are typically designed to generate convincing textures and cannot generalize well to multiple occluded scenarios simultaneously. By introducing bump mapping, we successfully added mid-level details to coarse 3D faces. More innovatively, our method takes into account occlusion scenarios. Thus on top of common 3D face reconstruction approaches, we in this paper propose a unified framework to handle multiple types of obstruction
    simultaneously (e.g., hair, palms and glasses \textit{et al.}).Extensive experiments and comparisons demonstrate that our method can generate high-quality reconstruction results with geometry details from captured facial images under occluded scenes.

\end{abstract}

\section{Introduction}
High-quality 3D face reconstruction is a fundamental problem in computer graphics~\cite{RN324} that is related to various applications such as digital animation~\cite{RN383}, video editing~\cite{RN383} and face recognition~\cite{zhang2022sttnet,zhang2022tmn}. Since Vetoer's first 3D face~\cite{RN823},3D reconstruction methods have rapidly advanced enabling applications. However, these methods all perform poorly in terms of face geometry details. To make the problem tractable, most proposed methods introduce existing statistical models or prior knowledge. These models are unable to reconstruct expression-dependent wrinkles, which are essential for analyzing human expression.
\\Several methods recover detailed facial geometry that lacks robustness to occlusions~\cite{RN766,RN215}. We introduce a novel face geometry detail generation method, which learns bump maps (simulate geometry changes) from in-the-wild face images with occlusion. In contrast to prior work (estimating mid-level features often breaks down), our method generates bump maps from a low-dimensional representation containing subject-speciﬁc detail parameters and expression parameters. Our detailed model builds upon this separation design. This design is fundamental, as it allows estimating a robust global shape, even under occluded scenes.
\\The main contributions are summarized as follows:
\\$\bullet$\ We propose a novel Face Image Synthesis Network, a simple yet effective diversity promoting face image regeneration approach. The regenerated eyeglasses removal face without glasses will guide the generation of a 3D model.
    \\$\bullet$\ We have improved the loss function of our 3D reconstruction system for occluded scenes with eyeglasses. Our results are more accurate than other approaches. As a result of our method, we are able to obtain state-of-the-art qualitative performance in real-world images.

\section{Related Work}
\subsection{Single image 3D face reconstruction}
Since the first 3DMM model was proposed by Blanz and Vetter~\cite{RN46}, single image based 3D face reconstruction has become a hot research topic and considerable progress have been made in the field. Richardson \textit{et al.}~\cite{RN121} presented a method based on CNN that can reconstruct 3D face based on synthetic data. As training deep neural networks usually demand a large amount of data to get acceptable results, Deng \textit{et al.}~\cite{RN239} proposed an approach that can achieve accurate 3D face reconstruction with weakly supervised learning based on less training data. Kemelmacher-Shlizerman and Basri~\cite{RN144} recovered 3D faces by exploiting the similarity of faces based on a single 3D reference model of a different person.Liu \textit{et al.}\cite{RN1359} built a 3D face model that can exploit both faces with fully labeled 3D landmarks and unlimited unlabeled in-the-wild face images. Lee \textit{et al.}~\cite{RN218} employed an uncertainty-aware encoder and a fully nonlinear decoder model for realistic 3D face reconstruction. Cheng \textit{et al.}~\cite{RN1360} solved the 3D face reconstruction problem based on graph convolutional networks obtaining good results without scarifying speed. Shang \textit{et al.}~\cite{RN767} proposed a self-supervised training architecture that is accurate and robust, even under large variations of expressions, poses, and illumination conditions. Li \textit{et al.}~\cite{RN1361} publicized an end-to-end framework and designed an efficient network model that can apparently increase the accuracy of face alignment and 3D face reconstruction. Li \textit{et al.}~\cite{RN1362} presented a multi-attribute regression reconstruction network that can work well in complex cases when provided with 2D images including severe poses, extreme expressions, and partial occlusions.
\subsection{Generative Adversarial Networks}
Generative adversarial networks (GANs) was first proposed by Goodfellow \textit{et al.} to study the generative model. Classical GANs consist of a generator and a discriminator. The aim of the generator is to generate data samples that can confuse the discriminator. The generator and the discriminator must improve themselves to win the 'game' until a Nash equilibrium is achieved; then generator successfully learns the distribution of the real dataset. GANs have been applied in many fields, including face image synthesis. Zhan \textit{et al.}~\cite{RN1363} proposed Spatial Fusion GAN (SF-GAN), which can obtain better results in both geometry and appearance spaces utilizing a geometry synthesizer and an appearance synthesizer. A triple-translation GAN (TTGAN) is proposed for face image synthesis by Ye \textit{et al.}~\cite{RN1364}. TTGAN adopts a triple translation consistency loss to translate from a rendered original input image to the desired output image. Sangloy \textit{et al.}~\cite{RN1365} proposed an adversarial image synthesis architecture that can extract information from sketched boundaries and parse color strokes and output realistic face images.
\subsection{Face image synthesis}
Deep pixel-level face generating has been studied for
several years. Many methods achieve remarkable results.
Context encoder~\cite{RN290} is the first deep learning network designed for image inpainting with encoder-decoder architecture. Nevertheless, the networks do a poor job in dealing with human faces. Following this work,
Yang \textit{et al.} used a modified
VGG network to improve the result of the
context-encoder, by minimizing the feature difference of
photo background. Dolhansky \textit{et al.} demonstrated the significance of exemplar data for inpainting. However, this method only focuses on filling in missing eye regions of the frontal face, so it does not generalize well.
EdgeConnect~\cite{RN353} shows impressive proceeds which disentangling generation into two stages: edge generator and image completion network. Contextual
Attention takes a similar two-step approach.
First, it produces a base estimate of the invisible region.
Next, the refinement block sharpens the photo by background patch sets. The typical limitations of current face image generate schemes are the necessity of manipulation, the complexity of fundamental architectures, the degradation in accuracy, and the inability of restricting modification to local region.
\section{Proposed Approach}
\begin{figure}[htb]
    \begin{center}
        \includegraphics[width=1.15\linewidth]{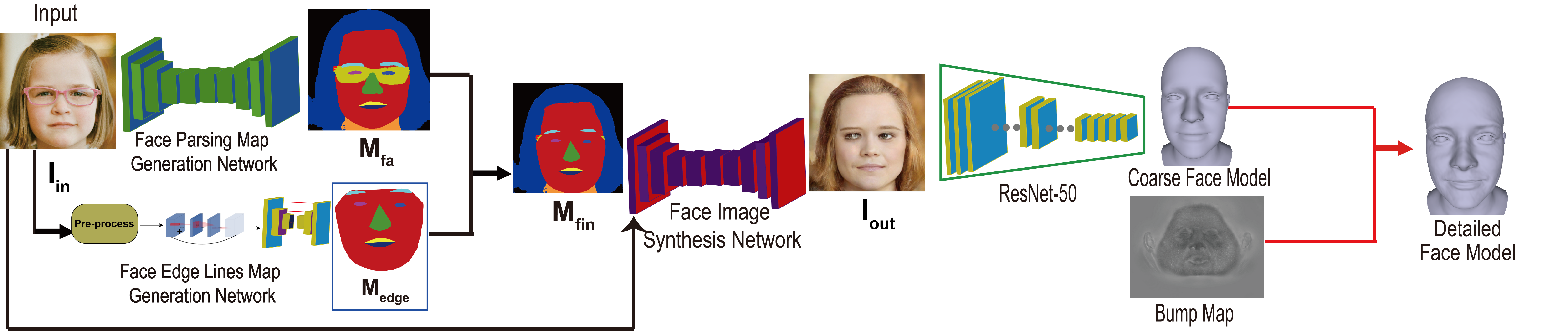}
    \end{center}
    \caption{{\textbf{Method overview.} At first, as input for our face image synthesis network, we need the target image ${{\mathbf{I}}_{\mathbf{in}}}$ and map ${{\mathbf{M}}_{\mathbf{fin}}}$. We utilize the face parsing map generation module and edge lines map generation module to obtain
                the map ${{\mathbf{M}}_{\mathbf{fa}}}$ and ${{\mathbf{M}}_{\mathbf{edge}}}$. Then we obtain the final face parsing map ${{\mathbf{M}}_{\mathbf{fin}}}$ following  Zhao \textit{et al.}'s Algorithm\cite{RN852}. After obtaining the face image ${{\mathbf{I}}_{\mathbf{out}}}$ with eyeglasses removed, in step two, we leverage ResNet-50 and texture refinement network to reconstruct the final 3D model.}}
    \label{fig:overall}
\end{figure}
We propose a detailed 3D face reconstruction method
(as shown in Figure~\ref{fig:overall}) based on a single photo that consists of
two steps:
\\$\bullet$\ in response to the occlusion area, synthesizing the 2D face with complete facial features.
    \\$\bullet$\ detailed 3D shape reconstruction module based on unobstructed frontal images.
\subsection{Face parsing map generation}
\begin{figure*}[ht]
    \centering
    \includegraphics[width=0.80\textwidth]{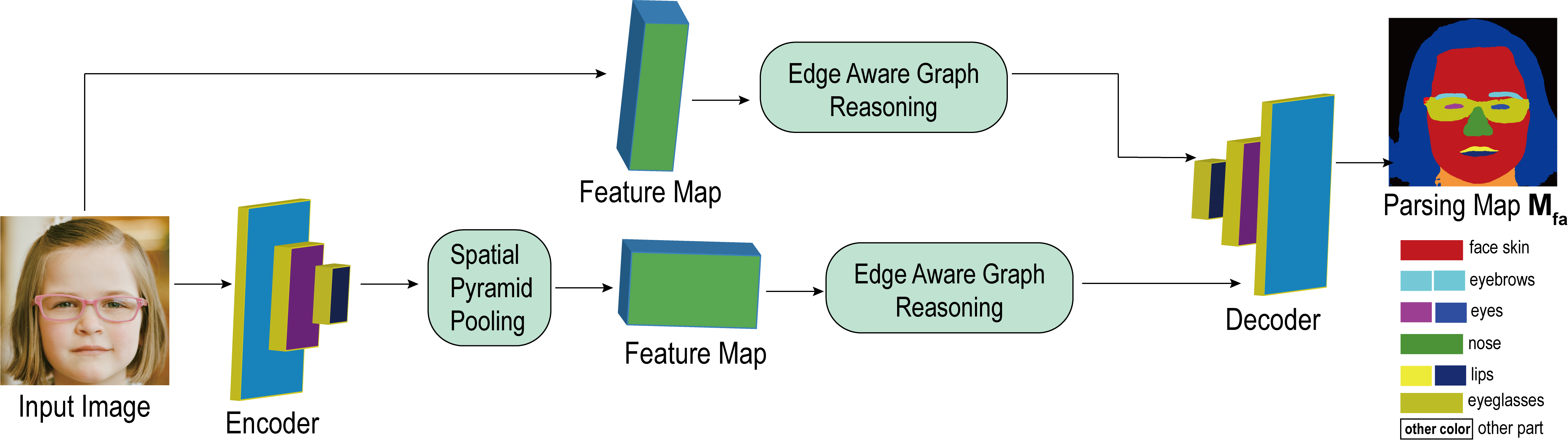}
    \caption{The overview of the proposed face parsing network.} \label{fig:02_face-parsing-network}
\end{figure*}
Our goal is to realize detailed 3D face shape reconstruction under occluded scenes using our method. Pixel-level recognition of eyeglasses areas serves as a key step for our framework to ensure accuracy. Face parsing is a fundamental facial analysis task. Recently, methods based on Fully Convolutional Networks have achieved remarkable results on this
task~\cite{RN1047,RN1048,RN1049}. As shown in Figure~\ref{fig:02_face-parsing-network}, given a squarely resized face
image ${{\mathbf{I}}_{\mathbf{in}}}\in {{\mathbb{R}}^{H\times W\times 3}}$, we aim to apply a modified encoder-decoder
network ${{\mathcal{N}}_{fa}}$ as the backbone frame for face parsing. We take ${{\mathcal{N}}_{fa}}$ to extract features at different levels for multi-scale illustration. In the structure
of ${{\mathcal{N}}_{fa}}$, high-level features contain semantic information while low-level features show local details, both of which are essential for face parsing. We feed the feature map with multi-scale information into the Edge Aware Graph Reasoning module, targeting to learn fundamental graph illustration for the characterization of the relations between vertices. The reasoning module consists of three components: graph projection operation, graph reasoning operation and graph reprojection operation. Let us make it clear. The graph projection operation projects the initial information onto vertices. The graph reasoning operation reasons the relational expression between regions over the graph and projects the acquired graph interpretation back to previous pixel grids. The graph reprojection operation leads to an optimized feature map with the same dimension and size. We implemented the reasoning module following the method of
Gusi \textit{et al.}~\cite{RN324}. Let us explain the last step of the network. We transmit the optimized features into a decoder to estimate the final pixel labels. In our network, two different level feature maps are concatenated into the decoder. The two feature maps are concatenated
by the $1\times 1$ convolution layer. The specific fusion method is through upsampling. That is, the high-level feature map is upsampled to the same dimension as the low-level feature map. Finally, we obtain the face parsing
map ${{\mathbf{M}}_{\mathbf{fa}}}\in {{\mathbb{R}}^{H\times W\times 1}}$.
\subsection{Face edge lines map generation}
\begin{figure*}[h]
    \centering
    \includegraphics[width=1.15\textwidth]{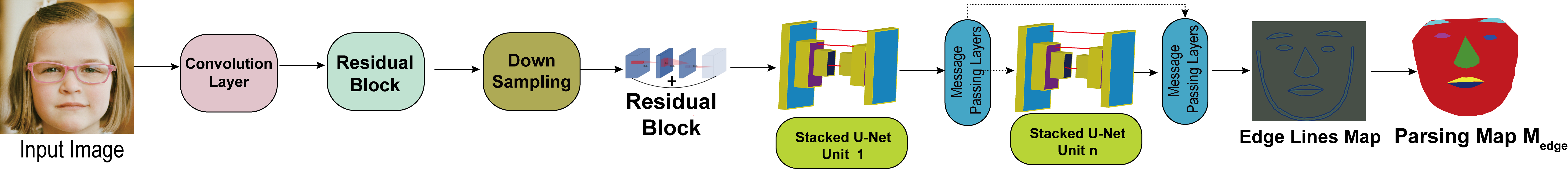}
    \caption{The overview of the proposed face edge lines map generation approach.} \label{fig:03_face-edge-lines-map}
\end{figure*}
In order to generate an accurate face parsing map, our method uses face edge lines to guide the reconstruction of the face parsing map. Face edge lines is closely related to the facial landmark. The reason why we choose face edge lines instead of landmarks is that landmarks have difficulties in presenting the accurate facial features
structure~\cite{RN544}. In this section, we describe the proposed face edge lines map generation framework in detail. As shown in Figure~\ref{fig:03_face-edge-lines-map} (a) and (b), the proposed framework consists of two parts: (a) face edge lines generation module; (b) face edge lines effectiveness discriminator.
\\As shown in Figure~\ref{fig:03_face-edge-lines-map} (a), stacked U-Nets is the core part of the face edge lines generation module. More than piecemeal landmarks, face edge lines can well describe the geometry structure of a face. Most of the previous convolutional networks only use the convolutional features of the last layer. Image information at other scales will be lost. Unlike the previous network, the main contribution of the stacked U-Nets
unit~\cite{RN1052,RN50} is to use multi-scale features to represent image information. We Leverage the mean squared error (MSE) between the estimated Face edge lines map and the ground-truth map. The presence of obstructions (this paper focuses on eyeglasses) will significantly affect the accuracy of edge lines generation. In order to relieve the loss of image information due to eyeglasses, we introduce message passing layers to pass information between face edge lines. It is proposed in this implementation that the feature map at the end of each stack should be divided
into $M$ (the number) areas. We implemented the message passing approach following the method of Chu \textit{et al.}~\cite{RN1046}. This process is visualized in
Figure~\ref{fig:03_face-edge-lines-map}.
\\\textbf{Intra-level message passing layer.} Among the steps involved in dealing with the problem of occlusion of eyeglasses, the intra-level message passing plays a crucial role. A layer such as this one is used at the end of each U-Nets stack in order to transmit information between visible edge lines and eyeglasses areas. Consequently, in the process of designing eyeglasses, the prediction of the eyeglasses areas can be improved through the visible edge lines data.
\\\textbf{Inter-level message passing layer.}It is true that there are various U-Nets stacks that focus on different dimensions of facial information, but in the case of multiple stacks, the facial information is transferred in the different stacks by performing communication between the former stacks and the latter stacks. When stacking more hourglass subnets, inter-level message passing is adopted to ensure that the face edge lines map maintains the quality when messages are passed from the lower stacks to the higher stacks.
\\\textbf{Adversarial learning for edge lines effectiveness.} Poor face edge lines map will adversely affect the accuracy of the 3D face model. When training, we use adversarial learning between the estimated edge lines map and the ground-truth map in order to guarantee the effectiveness of the edge lines map obtained in the generation stage. Using the Face edge lines map generator, the edge lines
map ${{\mathbf{M}}_{\mathbf{edge}}}\in {{\mathbb{R}}^{H\times W\times 1}}$ is generated with the coordinate
set ${{S}_{coor}}$; the mapping between the generated coordinate set and the ground-truth distance
matrix $\mathbf{M}{{\mathbf{A}}_{\mathbf{gt}}}$. In order to determine whether a generated edge line map is fake or not, the ground
truth ${{d}_{gt}}$ can be calculated as:
\begin{equation}
    {{d}_{gt}}({{\mathbf{M}}_{\mathbf{edge}}},{{S}_{coor}})=\left\{ \begin{aligned}
         & 0,Es{{t}_{s\in {{S}_{coor}}}}({{d}_{gt}}<\theta )<\delta \\
         & 1,\text{other cases}                                     \\
    \end{aligned} \right.
\end{equation}
where $Est$ denotes the probability value calculation function, $\theta $ denotes the distance threshold to ground truth edge
lines, $\delta $ denotes the probability threshold.
\\In order to combine the edge lines effectiveness discriminator $D$ and the face edge lines
map estimator $G$, we apply the concept of adversarial learning. The loss function of the discriminator $D$ can be calculated as:
\begin{equation}
    {{\mathcal{L}}_{D}}=\mathbb{E}[\log (1-\left| D(G({{\mathbf{I}}_{\mathbf{in}}}))-{{d}_{gt}} \right|)]-\mathbb{E}[\log D({{\mathbf{M}}_{\mathbf{gt}}})]
\end{equation}
where ${{\mathbf{M}}_{\mathbf{gt}}}$ denotes the ground truth face edge lines map. A discriminator is trained to predict an edge lines map on the ground truth as well as predict the generated edge lines map according
to ${{d}_{gt}}$ . With effectiveness discriminator, the adversarial loss can be calculated as:
\begin{equation}
    {{\mathcal{L}}_{adv-loss}}=\mathbb{E}\left[ \log (1-D(G({{\mathbf{I}}_{\mathbf{in}}})) \right]
\end{equation}
\subsection{Recovering 3D face geometric details}
We obtain the final face parsing map ${{\mathbf{M}}_{\mathbf{fin}}}$ following  Zhao \textit{et al.}'s Algorithm\cite{RN852}.
We synthesize the face photo ${{\mathbf{I}}_{\mathbf{out}}}$  by existing methods~\cite{RN311}.
Given ${{\mathbf{I}}_{\mathbf{out}}}$  , we used the ResNet to regress the corresponding coefficient $y$. Due to the collection of large scale high-resolution 3D texture datasets is still very costly and scarce, the ResNet was trained under weakly supervised. The corresponding loss function consists of
four parts~\cite{RN239,RN46}:
\begin{equation}
    {{\mathcal{L}}_{shape}}={{\lambda }_{feat}}{{\mathcal{L}}_{feat}}+{{\lambda }_{regu}}{{\mathcal{L}}_{regu}}+{{\lambda }_{phot}}{{\mathcal{L}}_{phot}}+{{\lambda }_{land}}{{\mathcal{L}}_{land}}
\end{equation}
Here we
set ${{\lambda }_{{fea}t}}{=0}{.2}$,${{\lambda }_{regu}}=3.6e-4$,${{\lambda }_{phot}}=1.4$,
${{\lambda }_{land}}=1.6e-3$  respectively in all our experiments.
\\The addition of human face geometric details is the core of our method. We choose to add a bump map on the
base shape ${{\mathbf{S}}_{\mathbf{basi}}}$.Inspired by the method of image-to-image translation
method, we define the displacements of the depth map as the distances through the pixels
of ${{\mathbf{I}}_{\mathbf{out}}}$ to the 3D face surface. Generally, we define the bump
map $\Phi (\mathbf{b})$ as:
\begin{equation}
    \Phi (\mathbf{b}){=}\left\{ \begin{aligned}
         & \phi (0)\ \ \ \text{other}\text{cases}                                                     \\
         & \phi ({d}'(\mathbf{b})-d(\mathbf{b}))\ \text{face}\ \text{projects}\ \text{to}\ \mathbf{b} \\
    \end{aligned} \right.
\end{equation}
where $\phi (\cdot )$ denotes an encoding function that converts the depth value to the linear
range $[0,…,255]$, $\mathbf{b}$ denotes the pixel coordinate $\left[ x,y \right]$ in ${{\mathbf{I}}_{\mathbf{out}}}$ , ${d}'(\mathbf{b})$ denotes the depth, which is the distance from the surface of the detailed face shape
to $\mathbf{b}$  along the line of sight, $d(\mathbf{b})$ denotes the depth of the basic shape.
\\Thus, Given a bump map $\Phi $  and the depth of the basic shape,we can compute the detailed depth
follows ${d}'(\mathbf{b}){=}d(\mathbf{b})+{{\phi }^{-1}}(\Phi (\mathbf{b}))$.In order to increase geometric details and to suppress noise, we define the loss function as
follows:
\begin{equation}
    {{\mathcal{L}}_{geo}}=\left\| \tilde{\Phi }-\Phi  \right\|+\left\| \frac{\partial \tilde{\Phi }}{\partial x}-\frac{\partial \Phi }{\partial x} \right\|+\left\| \frac{\partial \tilde{\Phi }}{\partial y}-\frac{\partial \Phi }{\partial y} \right\|
\end{equation}
where $\left\| \cdot  \right\|$ denotes the ${{L}_{1}}$ norm, $\tilde{\Phi }$ denotes the ground truth and $\frac{\partial \tilde{\Phi }}{\partial x}$ , $\frac{\partial \tilde{\Phi }}{\partial {y}}$ denotes the 2D gradient of the bump map.
After the 3D face is reconstructed, it can be projected
onto the image plane with the perspective projection:
\begin{equation}
    {{V}_{2d}}\left( \mathbf{P} \right)=f*{\mathbf{P}_{\mathbf{r}}}*{\mathbf{R}}*{\mathbf{S}_{\mathbf{mod} }}+{\mathbf{t}_{\mathbf{2d}}}
\end{equation}
where ${{V}_{2d}}\left( \mathbf{P} \right)$ denotes the projection
function
that turned the 3D model into 2D face
positions, $f$ denotes the scale factor,
${\mathbf{P}_{\mathbf{r}}}$ denotes the projection matrix,$\mathbf{R}\in SO(3)$ denotes the rotation matrix and ${\mathbf{t}_{\mathbf{2d}}}\in {{\mathbb{R}}^{3}}$  denotes the translation vector.
\\Therefore, we approximated the scene
illumination with Spherical Harmonics (SH)~\cite{RN642} parameterized by coefﬁcient
vector $\gamma \in {{\mathbb{R}}^{9}}$ .
In summary, the unknown parameters to be learned can be denoted
by a vector $y=({{\boldsymbol{\alpha }}_{\mathbf{id}}},{{\boldsymbol{\beta }}_{\mathbf{exp}}},{{\boldsymbol{\beta }}_{\mathbf{t}}},\boldsymbol{\gamma },\mathbf{p})\in {{\mathbb{R}}^{239}}$ ,where $\mathbf{p}\in {{\mathbb{R}}^{6}}=\{\mathbf{pitch},\mathbf{yaw},\mathbf{roll},f,{{\mathbf{t}}_{\mathbf{2D}}}\}$ denotes face poses. In this work, we used a fixed
ResNet-50 network to regress these coefficients.
\\We found that by adding these last two terms of loss function and we reduce bump map noise by favoring smoother surfaces. At the same time, the final effect shows that high-frequency details are preserved.
\section{Implementation Details}
All the networks were trained using the Adam solver~\cite{RN761}. To train our face parsing map generation network, we collected two sources dataset: Helen
dataset~\cite{RN520} and CelebAMask-HQ dataset~\cite{RN311}. The Helen dataset contains $2330$ images with $11$ categories: background, skin, paired lips, paired eyes, paired brows, paired mouth and hair. The CelebAMask-HQ dataset is a large-scale face parsing datasets which includes $30000$  high-resolution portrait images. The dataset contains $19$ categories. In addition to the facial unit, the components such as eyeglass, earring, necklace, neck, and cloth are also annotated.
\\In the face parsing map generation stage, our backbone is a modiﬁed version of the trained parsing
model~\cite{RN324}. We made the parsing model exclude the average pooling layer. For the pyramid pooling module, we follow the implementation of the method of Te \textit{et al.}~\cite{RN324} with exploiting global contextual information. We leveraged the fixed parsing model to
generate ${{\mathbf{M}}_{\mathbf{fa}}}$ . In the face edge lines map generation stage, all training images are cropped and resized
to $512\times 512$. We obtained ${{\mathbf{M}}_{\mathbf{edge}}}$  according the lines map generation network. We implemented message passing module following naturally obtains face features in different sizes. In the above two stages,we train our network on four datasets including
300W ($3148$ sample images)~\cite{RN743} and AFLW ($24386$ sample images)~\cite{RN1074}.
\section{Experimental Results}
In this work, we aim to generate a wide range of diverse and yet realistic 3D detailed reconstructions from occluded face images. Our approach should be characterized by the following three qualities: 1) the reconstructed geometry should fit as convincingly as possible to the visible regions, 2) the reconstructed model texture should not include eyeglasses, which is the essential requirement for the accuracy of the reconstruction.
\subsection{Qualitative comparisons with recent art}
\begin{figure*}[ht]
    \centering
    \includegraphics[width=0.60\textwidth]{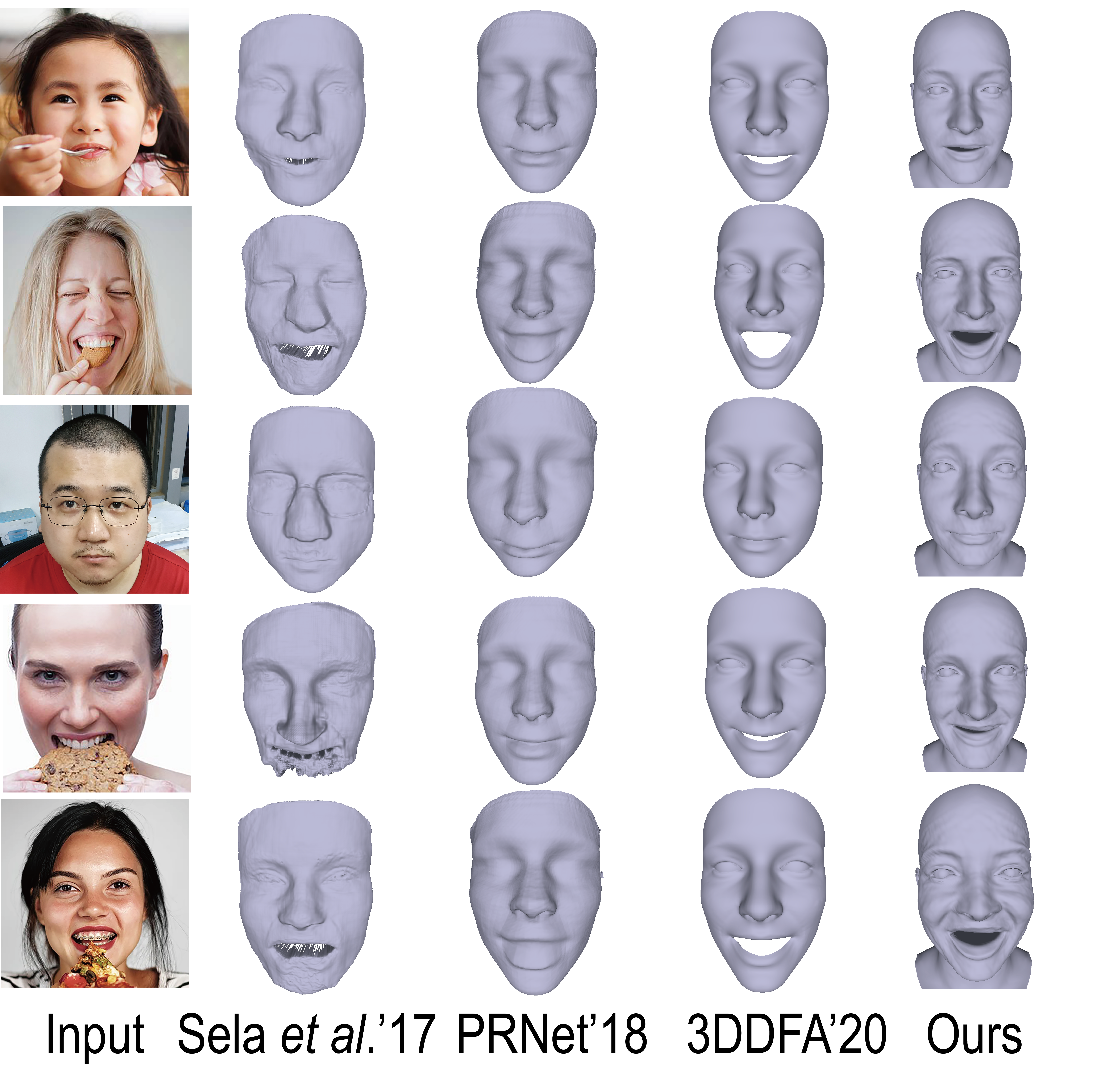}
    \caption{Comparison of qualitative results. Baseline methods from left to right: Sela \textit{et al.}, PRNet,3DDFA and our method.} \label{fig:07_dingxingjieguo}
\end{figure*}
Figure~\ref{fig:07_dingxingjieguo} shows our results compared with the other arts. The last columns show our results. The remaining columns demonstrate the results of
Sela \textit{et al.}~\cite{RN200},PRNet~\cite{RN400} and 3DDFA~\cite{RN186}.Our
results show that our results have better handled the occlusion area than other methods. Figure~\ref{fig:07_dingxingjieguo} shows that our method can reconstruct a complete face shape with geometry details under occlusion scenes such as glasses, food and fingers. The approach of 3DDFA was aimed at extremely large poses. Therefore, it cannot reconstruct a detailed face model under occluded scenes. Its shape lacks details. Other methods focused on generating high-resolution face textures instead of geometry details. At the same time, it must also be pointed out, the other methods cannot effectively deal with occluded scenes.
\subsection{Quantitative comparison with recent art}
\begin{figure}[htb]
    \centering
    \includegraphics[width=0.45\textwidth]{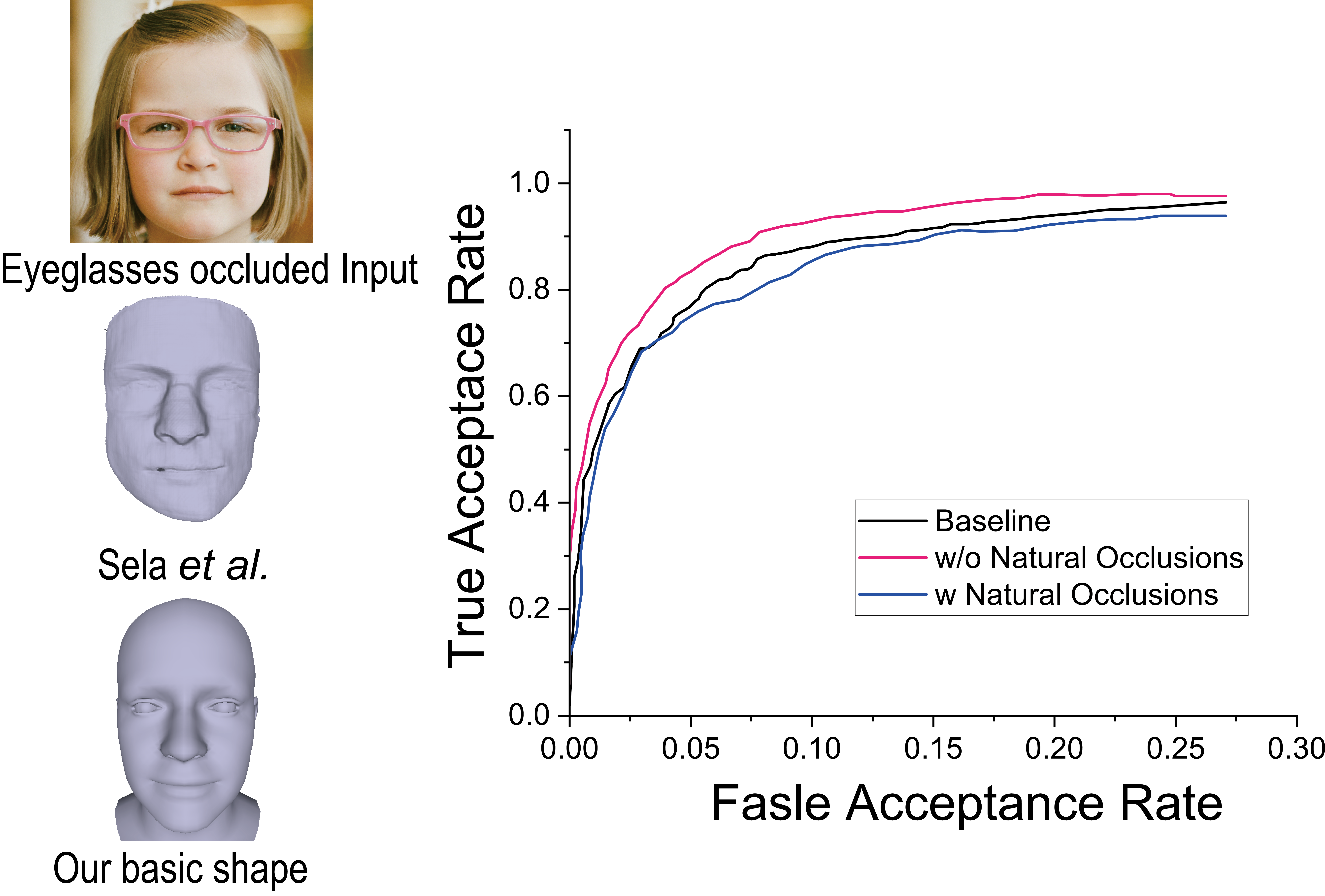}
    \caption{Reconstructions with eyeglasses. Left: Qualitative results of Sela \textit{et al.}~\cite{RN200} and our shape. Right: LFW veriﬁcation ROC for the shapes, with and without eyeglasses.} \label{fig:LFW}
\end{figure}
Our choice of using the ResNet-50 to regress the shape coefficients is motivated by the unique robustness to extreme viewing conditions in the paper of
Deng \textit{et al.}~\cite{RN239}. To fully support the application of our method to occluded face images, we test our system on the Labeled Faces in the Wild
datasets (LFW)~\cite{RN764} . We used the same face test system
from Anh \textit{et al.}~\cite{RN226}, and we refer to that paper for more details.
\\Figure~\ref{fig:LFW} (left) shows the sensitivity of the method of
Sela \textit{et al.}~\cite{RN200}. Their result clearly shows the outline of the eyeglasses. Their failure may be due to more focus on local details, which weakly regularizes the global shape. However, our method recognizes and regenerates the occluded area. Our method much robust provides a natural face shape under eyeglasses scenes. Though 3DMM also limits the details of shape, we use it only as a foundation and add refined texture separately.
\begin{table*}
    \centering
    \caption{Quantitative evaluations on LFW.}
    \label{biaoge:02}
    \begin{tabular}{p{0.9in}c c c c}
        \hline
        \multicolumn{1}{l}{Method}             & \multicolumn{1}{l}{100\%-EER} & \multicolumn{1}{l}{Accuracy} & nAUC           \\ \hline
        Tran \textit{et al.}~\cite{RN41}\qquad & $89.40\pm1.52$\qquad\qquad    & $89.36\pm1.25$\qquad\qquad   & $95.90\pm0.95$ \\
        Ours (w/ Gla)\qquad                    & $84.37\pm1.44$\qquad\qquad    & $85.79\pm0.42$\qquad\qquad   & $92.87\pm1.09$ \\
        Ours (w/o Gla)\qquad                   & $87.69\pm1.01$\qquad\qquad    & $89.02\pm0.89$\qquad\qquad   & $95.37\pm0.65$ \\ \hline
    \end{tabular}
\end{table*}
\\We further quantitatively verify the robustness of our method to eyeglasses. Table~\ref{biaoge:02} (top) reports veriﬁcation results on the LFW benchmark with and without eyeglasses (see also ROC in Figure~\ref{fig:LFW}-right). Though eyeglasses clearly impact recognition, this drop of the curve is limited, demonstrating the robustness of our method.
\section{Conclusions}
In this work, we describe a 3D face detailed reconstruction framework that can run efficiently under occluded scenes. Our method enables unobstructed face image synthesis by concatenating the original face parsing map with the face edge lines map which both are extracted from the input face image in the encoder-decoder network. The experiments on 3D face reconstruction using various datasets have shown that our method can effectively remove eyeglasses with equivalent quality and better accuracy control than the existing methods.
\\
\\\textbf{Acknowledgements}
This paper is supported by National Natural Science Foundation of China (No. 62072020) and the Leading Talents in Innovation and Entrepreneurship of Qingdao (19-3-2-21-zhc).
%
%
%

%




%
%
\bibliographystyle{splncs04}
\bibliography{ref.bib}
\end{document}